\documentclass[runningheads,a4paper]{llncs}

\usepackage{times}
\usepackage{amssymb}
\usepackage{graphicx}
\usepackage{url}
\usepackage[normalem]{ulem}

\usepackage[disable]{todonotes} 
\newcommand{\todoin}[2][]{\todo[inline,#1]{#2}}
\newcommand{\todoinb}[2][]{\todo[inline,color=blue!40,#1]{#2}}
\newcommand{\todoing}[2][]{\todo[inline,color=green!40,#1]{#2}}
\def\rrr#1{{\color{red}#1}}
\def\ddd#1{{\color{blue}#1}}

\def\tm#1{{\color{purple}TM: #1}}
\def\jl#1{{\color{magenta}JL: #1}}

\usepackage[utf8]{inputenc}
\usepackage{overpic}

\newcommand{\keywords}[1]{\par\addvspace\baselineskip
\noindent\keywordname\enspace\ignorespaces#1}

\urldef{\mailsa}\path|musil@ufal.mff.cuni.cz|
\urldef{\mailsb}\path|author3@institute2.org|

\begin{document}

\title{Examining Structure of Word Embeddings with PCA}

\titlerunning{Examining Structure of Word Embeddings with PCA}

\author{Tom\'{a}\v{s} Musil}


\authorrunning{Tom\'{a}\v{s} Musil}

\institute{Charles University, Faculty of Mathematics and Physics,
Institute of Formal and Applied Linguistics,
Malostranské náměstí 25, 118 00 Prague, Czech Republic \\
\url{http://ufal.mff.cuni.cz/tomas-musil} \\
\mailsa\\
}


\toctitle{} \tocauthor{}

\maketitle

%
%
%
%
\begin{abstract}
In this paper we compare structure of Czech word embeddings for English-Czech neural machine translation (NMT), word2vec and sentiment analysis. We show that although it is possible to successfully predict part of speech (POS) tags from word embeddings of word2vec and various translation models, not all of the embedding spaces show the same structure. The information about POS is present in word2vec embeddings, but the high degree of organization by POS in the NMT decoder suggests that this information is more important for machine translation and therefore the NMT model represents it in more direct way. Our method is based on correlation of principal component analysis (PCA) dimensions with categorical linguistic data. We also show that further examining histograms of classes along the principal component is important to understand the structure of representation of information in embeddings.

\end{abstract}

\section{Introduction}

Embeddings of linguistic units (sentences, words, subwords, characters, \ldots{}) are mappings from the discrete and sparse space of individual units into a continuous and dense vector space. Since neural networks became the dominant machine learning method in NLP, embeddings are used in majority of NLP tasks.

Embeddings can be either learned together with the rest of the neural network for the specific task, or pretrained in a neural network for different task, typically language modelling. Because different information is useful in various tasks, the embeddings may vary with respect to the information they represent.

In this paper, we examine the structure of word embeddings for Czech words.  We compare the word2vec model embeddings with embeddings from three different neural machine translation architectures (for translation between Czech and English) and embeddings from a convolutional neural network for sentiment analysis. We show how these models differ in the structure of information represented in the embeddings. \todoin{\jl{tady by to chtělo vysvětli, proč sis vybral zrovna tyhlety tasky}\tm{protože byly zrovna po ruce a někde člověk začít musí, ale to mi nepřipadá zajímavé sem psát}\rrr{RR: můžeš ocitovat slona a říct že proto že na to jsou modely pro češtinu}}

In Section~\ref{sec:rel} we discuss related work. In Section~\ref{sec:tasks} we describe the tasks and the models that we are solving them with. To compare the embeddings from different tasks, we show what information is contained in embeddings in Section~\ref{sec:cls}. In Section~\ref{sec:structure} we examine the structure of the vector space with respect to the represented information. We summarize the findings in the concluding Section~\ref{sec:concl}.

\section{Related Work}
\label{sec:rel}

A comprehensive survey of word embeddings evaluation methods was compiled by Bakarov \cite{bakarov2018survey}. An overview can also be found in the survey of methodology for analysis of deep learning models for NLP by  Belinkov \cite{belinkov2018analysis}.

Hollis and Westbury \cite{hollis2016principals} extracted semantic dimensions from the word2vec embeddings. They used principal component analysis as candidate dimensions and measured their correlation with various psycholinguistic quantities. \todoin{\jl{na první čtení není jasný, co jsou semantic dimensions}}
Our approach is different in that we examine categorical discrete information in word embeddings.
\todoing{and describe the structure }

Quian et al. \cite{qian2016investigating} investigate properties (including POS and sentiment score) of embeddings from 3 different language model architectures for more than 20 languages including Czech. They use a multilayer perceptron classifier trained on a part of the vocabulary and evaluated on the rest.

Saphra and Lopez \cite{W18-5438} show that language models learn POS first. They use singular vector canonical correlation analysis to “demonstrate that different aspects of linguistic structure  are  learned  at  different  rates,  with  part of speech tagging acquired early and global topic information learned continuously”.

Belinkov et al. \cite{belinkov2017neural,belinkov2018evaluating} evaluate what NMT models learn about morphology and semantics by training POS, morphological, and semantic taggers on representations from the models. \todoin{What do Neural Machine Translation Models Learn about Morphology? Effect of attention}
Chen et al. \cite{chen2013expressive} propose several tasks that will help us better understand the information encoded in word embeddings, one of them being sentiment polarity.
Our approach differs from these in that we are examining the structure of the embeddings, not only predicting the information from them. 

\section{Tasks, Models and Data}
\label{sec:tasks}

\def\rnnmodel{{\scshape RNN model}}
\def\rnnenc{{\scshape RNN NMT encoder}}
\def\rnndec{{\scshape RNN NMT decoder}}
\def\attmodel{{\scshape attention model}}
\def\attenc{{\scshape \attmodel{} encoder}}
\def\attdec{{\scshape \attmodel{} decoder}}
\def\transmodel{{\scshape transformer model}}
\def\transenc{{\scshape \transmodel{} encoder}}
\def\transdec{{\scshape \transmodel{} decoder}}
\def\wordtovec{{\scshape word2vec}}

In this section we describe the tasks for which the examined models were trained, the models themselves and the data we used to train them.

\subsection{Neural Machine Translation}

Neural machine translation (NMT) started with the sequence-to-sequence recurrent neural network architecture \cite{sutskever2014sequence}. The first model to significantly surpass the previous paradigm of statistical MT introduced the attention mechanism  \cite{bahdanau2014neural}. Further development proved that attention is all you need 
 \cite{vaswani2017attention}.

We trained our NMT models on the CzEng \cite{czeng16:2016} Czech-English parallel corpus, the \emph{c-fiction} section. The models were trained with the \emph{Neural Monkey} toolkit \cite{NeuralMonkey:2017}. The embedding size was 512, vocabulary size 25k. The embeddings were taken from the Czech part of the MT system, that is from the \emph{cs-en} encoder and \emph{en-cs} decoder. We compare three different models:

The \rnnmodel{} is a LSTM encoder and decoder with a hidden state of size 1024.

The \attmodel{} is bidirectional conditional GRU \cite{firat2016cgru} with attention \cite{bahdanau2014neural}.

The \transmodel{} is the \emph{Neural Monkey} implementation of the transformer architecture \cite{vaswani2017attention}, with 6 layers and 16 attention heads in each layer.

\subsection{Neural Language Models}

Neural language models used word embeddings before machine translation and other applications \cite{bengio2003neural}. Eventually they inspired the word2vec/Skip-gram approach \cite{word2vec,mikolov2013distributed}, which can be thought of as an inverted language model: it is trained by predicting the context words from a word that is given.

For the word2vec embeddings to be comparable with the ones from NMT models, we trained the word2vec model on the Czech side of the parallel corpus that we used for the translation models. We used the Skip-gram model with embedding size 512, window size 11, and negative sampling with 10 samples.

\subsection{Sentiment Analysis}

Sentiment analysis is a task of deciding whether a given text is positive, negative, or neutral in its emotional charge. It find applications in marketing and public relations. Comments of products or services accompanied by a rating are a valuable data resource for this task.

For training the sentiment analysis models we used the CSFD CZ dataset \cite{habernal2013sentiment}, where data were obtained from user comments in the ČSFD film database.\footnote{\url{https://www.csfd.cz/}}
The model is a convolutional neural network based on \cite{kim2014convolutional} with embedding size 300, kernels of size 3, 4 and 5 with output dimension 100, dropout 0.8, and a classifier with 100 hidden units.
The model was trained with batch size 500, for 40 epochs.

\subsection{Part of Speech Tagging}


Part of speech tags for Czech words were obtained from MorfFlex \cite{morff}. It contains morphologically analyzed wordlist for Czech. The POS is the first position in MorfFlex morphological tag and there are 10 possible values for Czech POS (and two possible values for special cases that we did not use): 
Adjective (A),	
Numeral (C),	
Adverb (D),	
Interjection (I),	
Conjunction (J),	
Noun (N),	
Pronoun (P),	
Verb (V),	
Preposition (R),	
and Particle (T).
Some word forms are homonymous with other words with different POS. In that case, we mark that word as multivalent.
\todoinb{čeština má pevně POS, minimum homonymie; má to být tady, nebo až v další sekci?}

\todoin{\ddd{Pridal bych velikost slovniku a informaci, ze spousta je toho vygenerovanyho, ale ty to evaluujes stejne jenom na pruniku.}}

\section{What Information is Represented in Word Embeddings}
\label{sec:cls}

To show that some information is represented in word embeddings, we can train a model to predict that information an embedding alone (with no context). If the model correctly predicts the information for previously unseen embeddings, then the information must be inferable from the embeddings.

We evaluate the model's ability to learn to predict the information by crossvalidation. We divide the vocabulary randomly into 10 bins and train the model 10 times, each time leaving 1 bin for evaluation and training on the remaining 9.

\subsection{Part of Speech Classification}
\label{sec:pos_cls}

We use an intersection of the vocabulary of our translation models (25k word forms) with the MorfFlex dictionary and discard word forms occuring with more than one different POS tags in the dictionary. We are left with 21\,882 forms.

For predicting the POS tags, we use a simple multilayer perceptron classifier, with one hidden layer of size 100 and a softmax classification layer.

\begin{table}
\centering
\begin{tabular}{lc}
model & accuracy \\
\hline
\rnnenc{} & 94.69\,\% (+/- 0.93\,\%) \\
\rnndec{} & \textbf{97.77\,\%} (+/- 1.16\,\%) \\
\attenc{} & 96.17\,\% (+/- 1.39\,\%) \\
\attdec{} & 96.12\,\% (+/- 1.27\,\%) \\
\transenc{} & 96.37\,\% (+/- 1.49\,\%) \\
\transdec{} & 93.36\,\% (+/- 3.86\,\%) \\
\wordtovec{} & 95.01\,\% (+/- 1.94\,\%) \\
\end{tabular}
\caption{Accuracy of predicting POS classes from word embeddings taken from various models. The number in parenthesis is twice the standard deviation, covering the 95\,\% confidence interval.}
\label{tab:pos_klas}
\end{table}

In Table~\ref{tab:pos_klas}, we compare classification accuracies of POS tag predictions from word embeddings learned in different network architectures described in Section~\ref{sec:tasks}. For each NMT architecture we have trained models for both translation directions. We are examining the embeddings from the Czech side of the model, that is encoder for Czech-English translation and decoder for English-Czech translation.

We observe that in case of \rnnmodel{} without attention mechanism, POS tags can be significantly better predicted from decoder embeddings than from the encoder embeddings. The reason may be the fact that the decoder is trained to produce grammatically correct sentences and therefore needs to store this kind of information about words. The encoder, on the other hand, only needs to be aware of POS tags to the extent of its contribution to meaning disambiguation.

The \attmodel{} performs slightly worse, which may be because this model has additional parameters in the attention mechanism to store information in and therefore does not need to represent it in the embeddings.

The information may be even more scattered in case of the \transmodel{}, which uses multiple attention layers. That may explain why the results for the decoder have the highest variance. The lower performance compared to encoder embeddings classifier may seem to contradict our previous hypothesis (that decoder needs more information about morphology), but due to the high variance this difference is not significant.

The \wordtovec{} model is trained on the Czech side of the parallel data on which the NMT models were trained.

There are no major differences in the classifier performance on embeddings from different models.  Since it performs well on all of these embeddings, it may seem that POS is of roughly same importance for all of the models. However, in the next section we will show there are significant differences.

\section{Structure of the Vector Space}
\label{sec:structure}

In Section~\ref{sec:cls} we have seen that there is not much difference in the amount of the examined information represented in embeddings from different models. However, there may be difference in the structure of the representation.

One major obstacle in examining the structure of the embedding space is the random initialization, which means that two runs of the same experiment may render completely different embedding spaces and a single dimension or direction in the embedding space has no fixed interpretation.

Inspired by \cite{hollis2016principals}, we use principal component analysis 
and examine the correlation of the information in question with individual components. 
This helps us to establish invariants and visualise the differences in the structure. Principal component analysis (PCA) is a transformation that is defined in a way that the first component has the highest variance. The next component is the direction that has the highest variance and is ortogonal to the first component, and so on.

When we find a principal component that higly correlates with information that interests us, we can further examine the structure of the representation by plotting the information along the principal component.

\subsection{Emebedding Space and Part of Speech}

\begin{figure}
    \includegraphics[width=0.48\textwidth]{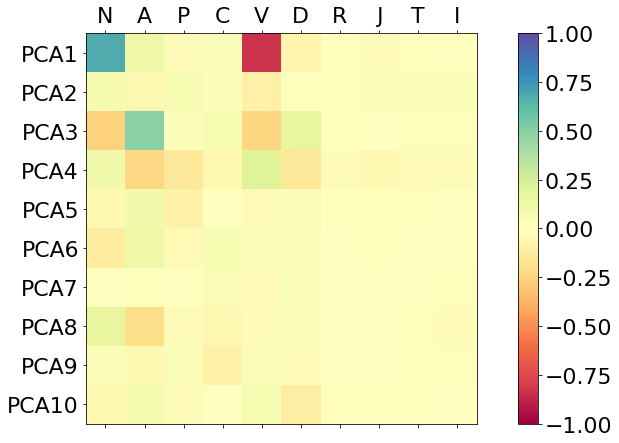}\hfill
    \includegraphics[width=0.48\textwidth]{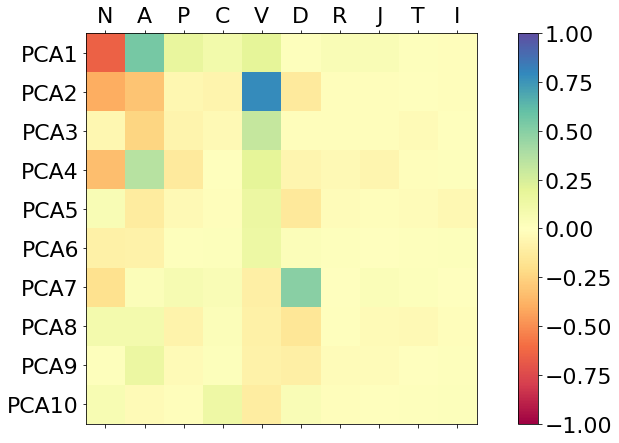}
    \newline\hbox{} \hfill \rnnenc{} \hfill \hfill \rnndec{} \hfill \hbox{}
    \newline\includegraphics[width=0.48\textwidth]{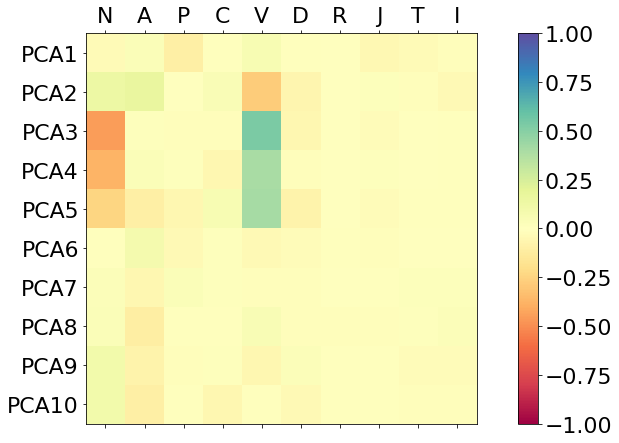}\hfill\begin{tabular}[b]{c c}
        N & Noun\\
        A & Adjective\\
        P & Pronoun\\
        C & Numeral\\
        V & Verb\\
        ~ & ~\\
        ~ & ~\\
        ~ & ~\\
        \end{tabular}\hfill\begin{tabular}[b]{c c}
        D & Adverb\\
        R & Preposition\\
        J & Conjunction\\
        T & Particle \\
        I & Interjection\\
        ~ & ~\\
        ~ & ~\\
        ~ & ~\\
    \end{tabular}\hfill\hbox{}
    \newline\hbox{} \hfill word2vec \hfill \hfill ~ \hfill \hbox{}
    
    \caption{Correlations of POS and PCA dimensions from the encoder of the Czech-English RNN NMT model (top left), the decoder of the English-Czech RNN NMT model (top right) and the word2vec model (bottom). The direction of the PCA dimensions is arbitrary, so the sign of the correlation is not important in itself, only if there are values with opposite signs in the same row we know that they are negatively correlated.
    }
    \label{fig:corr_enko}
\end{figure}

We measure a correlation between the principal components of the embeddings and binary vectors indicating whether a word belongs to a given POS category.
For each of the POS categories, we generate an indication vector with a position for each word from the vocabulary, containing 1 if that word falls under the particular category and 0 otherwise. We then measure correlation between these vectors and principal components of the embeddings.

\begin{figure}
  \centering
    \hbox{}\hfill \rnnenc{} \hfill \hfill \rnndec{} \hfill\hbox{}\newline
    \includegraphics[width=.48\textwidth]{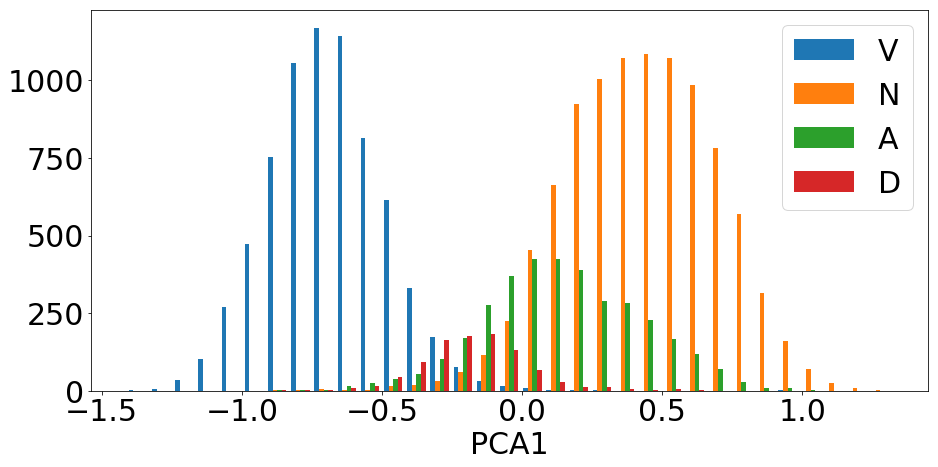}
    \includegraphics[width=.48\textwidth]{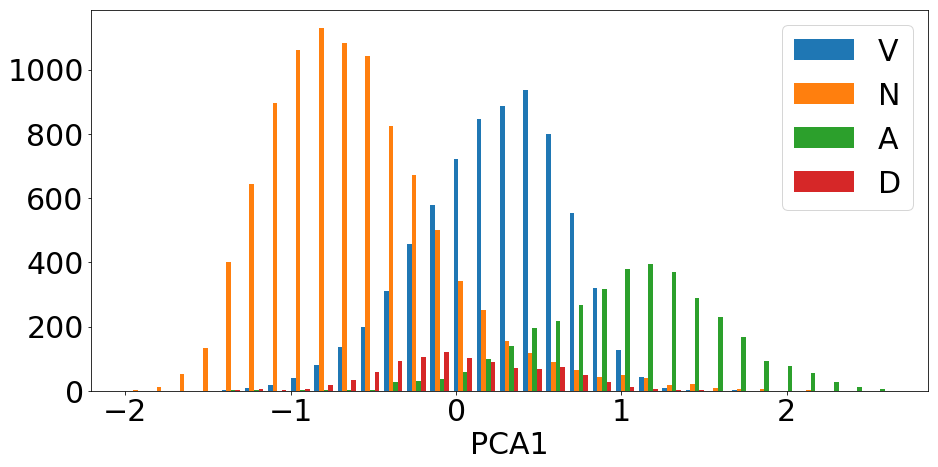}
    \includegraphics[width=.48\textwidth]{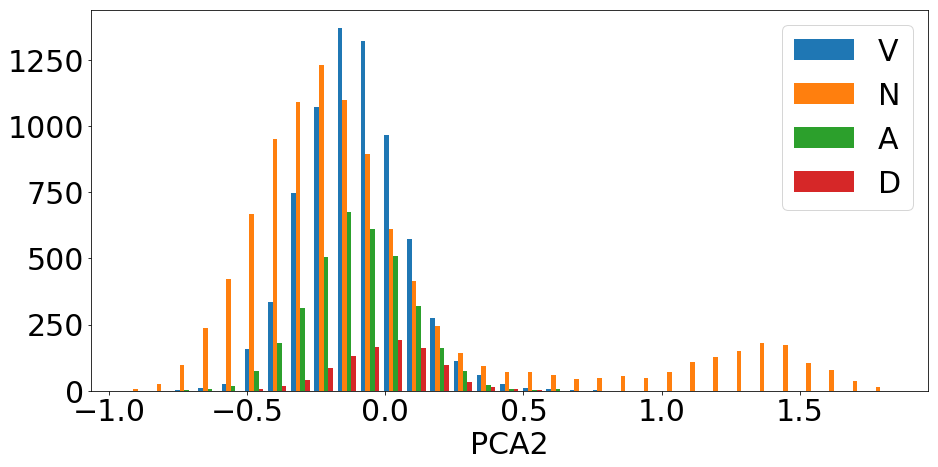}
    \includegraphics[width=.48\textwidth]{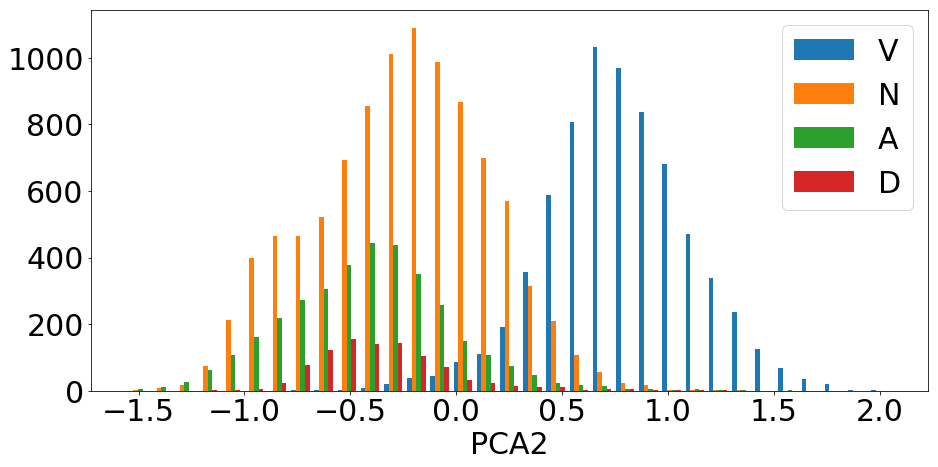}
    \includegraphics[width=.48\textwidth]{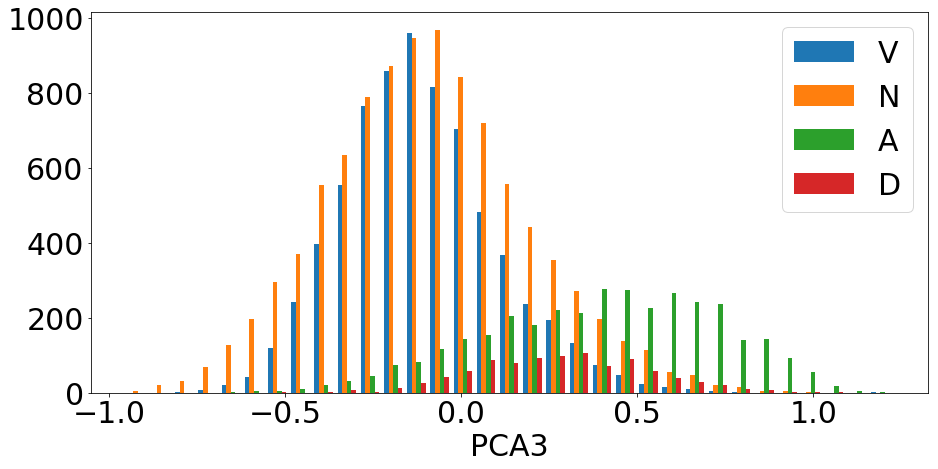}
    \includegraphics[width=.48\textwidth]{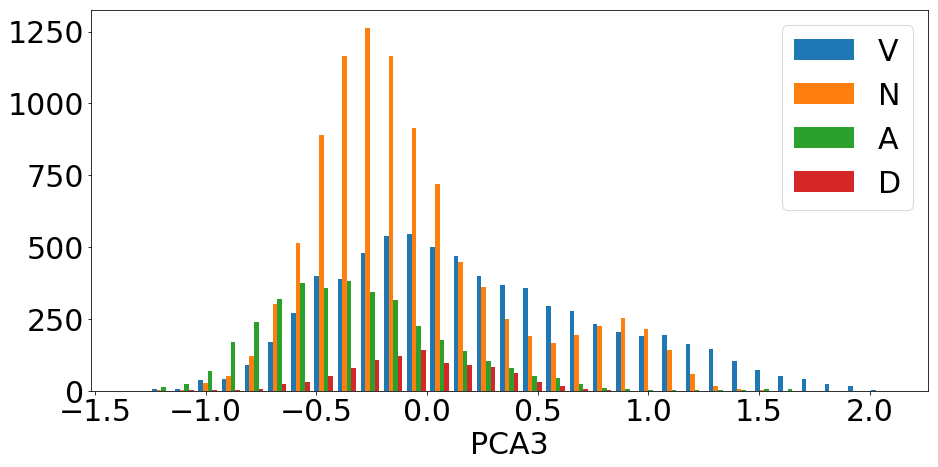}
    \caption{Histograms of the four largest POS classes along the first three PCA dimensions of the embeddings from the NMT \rnnmodel{}. The Czech-English \rnnenc{} is on the left and the English-Czech \rnndec{} on the right. V = verbs, N = nouns, A = adjectives, D = adverbs.}
    \label{fig:vnady}
\end{figure}

For example, in the correlation matrix for embeddings from the NMT RNN encoder in Figure~\ref{fig:corr_enko} we see a strong correlation between the first principal component and verbs. This correlation is negative, but that does not matter, because the direction of the principal components is arbitrary. We also see a strong correlation between the first principal component and nouns. In this case, the correlation is positive. That by itself is again arbitrary, but it is important that the correlation with verbs has the opposite sign. It means that in the direction of the first principal component verbs are concentrating on one side of the embedding space and nouns on the other. And indeed when we plot a histogram of the POS categories along the first principal component in Figure~\ref{fig:vnady}, we see that is the case.

As we have demonstrated in Section~\ref{sec:cls}, the NMT RNN architecture without attention stores more information about POS categories in the embeddings than architectures with attention, so we will examine the structure of its embedding space.

\begin{figure}
    \includegraphics[width=0.33\textwidth]{korela_enkoder.png}
    \includegraphics[width=0.33\textwidth]{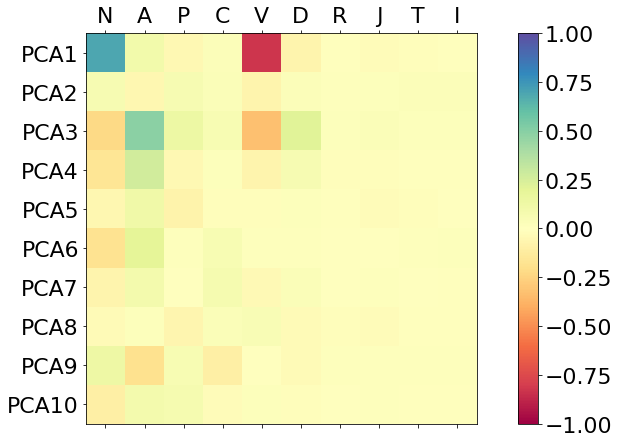}
    \includegraphics[width=0.33\textwidth]{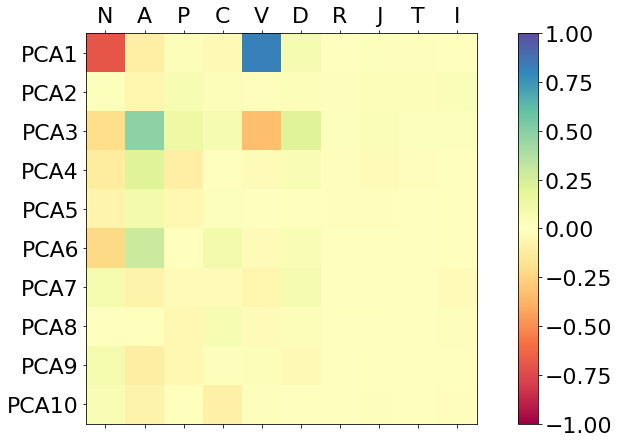}
    \caption{Correlations of POS and PCA dimensions of the \rnnenc{} embeddings from three runs of the Czech-English \rnnenc{} with different random initialization.  The direction of the PCA dimensions is arbitrary, so inversion of the first PCA dimension in the third matrix does not matter
    (see Figure~\ref{fig:corr_enko} for more information).
    }
    \label{fig:corr_enko_runs}
\end{figure}

There is an obvious difference in POS representation between encoder and decoder. This is not an artifact of random initialization. After we apply the PCA, we see the same structure across models with different random seeds, as in Figure~\ref{fig:corr_enko_runs}.

When we compare the correlation matrices for the encoder and the decoder embeddings in Figure~\ref{fig:corr_enko}, we see that in the encoder:
\begin{itemize}
\item verbs and nouns are strongly distinguished in the first dimension of the PCA,
\item there are adjectives (and adverbs to a lesser extent) on one side and nouns with verbs on the other in the third dimension,
\end{itemize}
whereas in the decoder:
\begin{itemize}
\item the first dimension of the PCA distinguishes adjectives from nouns,
\item the second dimension distinguishes verbs from nouns and adjectives,
\item the seventh dimension correlates with adverbs,
\item and overall the correlation with POS classes seems to be more pronounced than in the encoder. 
\end{itemize}

To better understand the structure of the POS representation in the embedding space we can look at the histograms of the four most important POS classes along the dimensions of the PCA of the embeddings. The histogram for the first dimension of the \rnnenc{} embeddings is plotted in Figure~\ref{fig:vnady}. It demonstrates in this dimension verbs are separated from the rest of the POS categories.

For the first PCA dimension of the \rnndec{} embeddings in Figure~\ref{fig:vnady} we see that nouns are concentrated on one side, adjectives on the other side and verbs with adverbs are in the middle. 

For the second PCA dimension of the \rnndec{} embeddings in Figure~\ref{fig:vnady} we see that verbs are concentrated on one side and all other categories on the other side. 

\begin{figure}
  \centering
    \hbox{}\hfill \rnnenc{} \hfill \hfill \rnndec{} \hfill\hbox{}\newline
    \includegraphics[width=.48\textwidth]{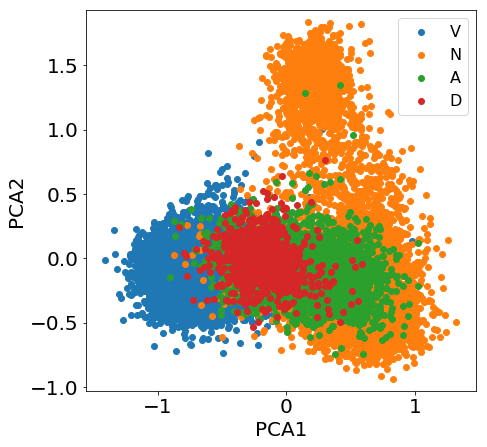}
    \includegraphics[width=.46\textwidth]{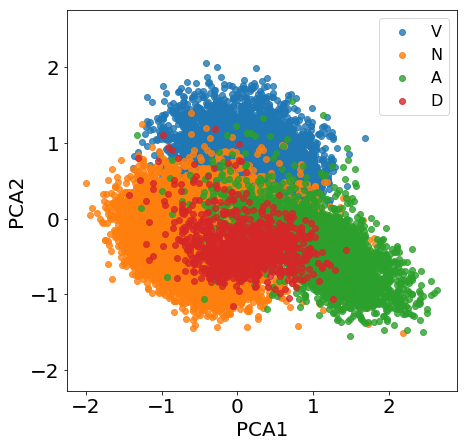}
    \newline\hbox{}\hfill \rnnenc{} \hfill \hfill \rnndec{} \hfill\hbox{}\newline
    \includegraphics[width=.48\textwidth]{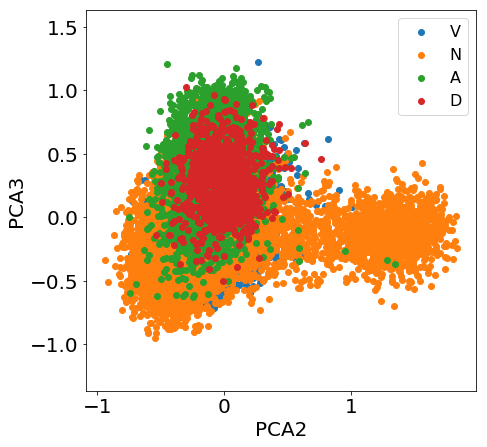}
    \includegraphics[width=.48\textwidth]{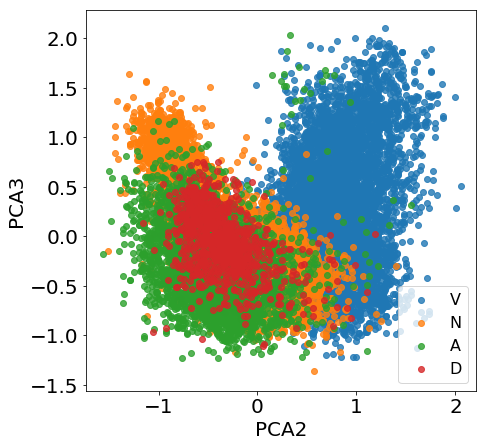}
    \caption{Distribution of the four largest POS classes along the first/second and second/third PCA dimensions of the embeddings from the the NMT \rnnmodel{}. The Czech-English \rnnenc{} is on the left and the English-Czech \rnndec{} on the right. V = verbs, N = nouns, A = adjectives, D = adverbs.}
    \label{fig:enk2d}
\end{figure}

The second PCA dimension shows interesting distribution of nouns in Figure~\ref{fig:vnady}. There is a separate cluster of nouns, which is even more apparent if we plot the distribution along two PCA dimensions in a planar graph in Figure~\ref{fig:enk2d}. When we take a sample of words from this cluster, it contains almost exclusively named entities: \emph{Fang, Eliáši, Još, Aenea, Bush, Eddie, Zlatoluna, Gordon, Bellondová, and Hermiona}.

There is a similar distribution of nouns in the third PCA dimension of the \rnndec{} embeddings. The smaller group again consists of named entities.

\begin{figure}
  \centering
    \includegraphics[width=.32\textwidth]{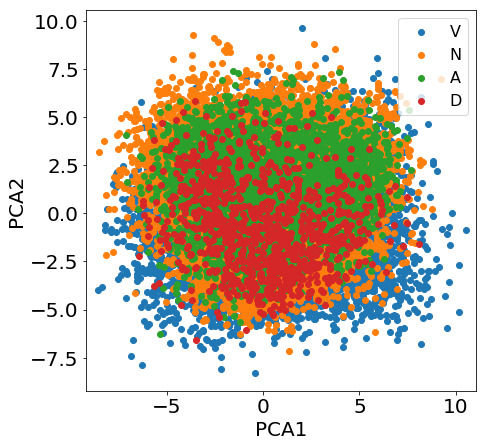}
    \includegraphics[width=.32\textwidth]{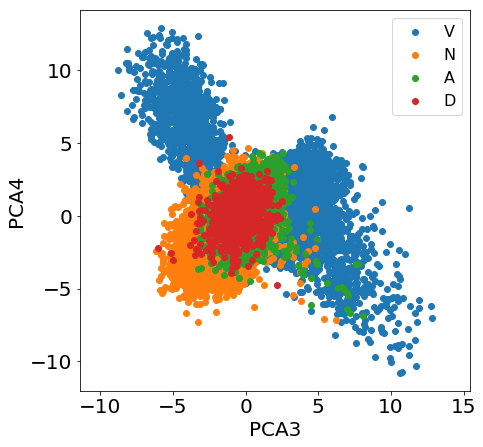}
    \includegraphics[width=.32\textwidth]{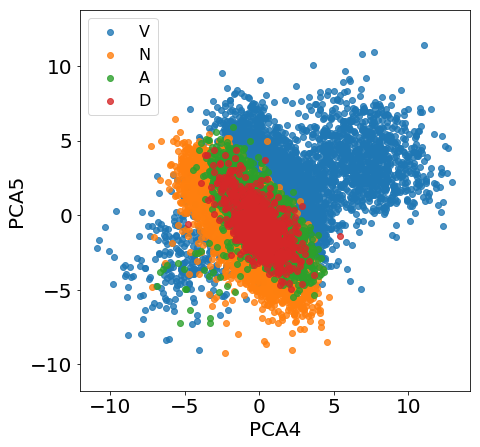}
    \caption{Distribution of the four largest POS classes along the first/second (left), the third/fourth (middle), and the fourth/fifth PCA dimensions of the embeddings from the word2vec model. V = verbs, N = nouns, A = adjectives, D = adverbs.}
    \label{fig:w2v2d}
\end{figure}

As we can already tell from Figure~\ref{fig:corr_enko}, the analysis of the \wordtovec{} embeddings is different. With respect to POS, the embeddings look normally distributed in the first dimensions of the PCA (see also the left part of Figure~\ref{fig:w2v2d}). There is a correlation with nouns and an inverse correlation with verbs in dimension three to five. In the middle and right parts of Figure~\ref{fig:w2v2d}, we see that verbs are composed of several clusters. When we take a sample from the cluster that is in the top left of the PCA3/PCA4 plot or in the top right of the PCA4/PCA5 plot, it concains almost exclusively infinitive verb forms: \emph{odpovědět, přenést, zabránit, získat, říct, předvídat, přiznat, opřít, přemoci, setřást}. The other protrusion of verbs in the lower right of the PCA3/PCA4 plot and lower left of the PCA4/PCA5 plot contains modal verbs: \emph{mohli, mohl, lze, nemohla, měla, musel, museli, musí, dokázala, podařilo}.

This is a stronger result than just being able to predict these categories with a classifier: not only can the POS (named entities, verb forms and possibly other categories) be inferred from the embeddings, but the embeddings space is structured according to these categories.

\subsection{Embedding Space and Sentiment}

In this section, we examine the word embeddings from the sentiment analysis model. When we computed the PCA correlation matrix for POS classes just like in the previous section, we found that there is no significant correlation (the correlation coefficient with the largest absolute value is 0.0680, more than 10 times lower than in the previous section, although the vocabulary is different, therefore the numbers are not directly comparable). The same classifier that predicts POS with more than 93\,\% accuracy from the translation model embeddings did not converge at all at the embeddings from this task. However, there is a different interesting structure.

\begin{figure}
  \centering
  \begin{overpic}[width=1\textwidth]{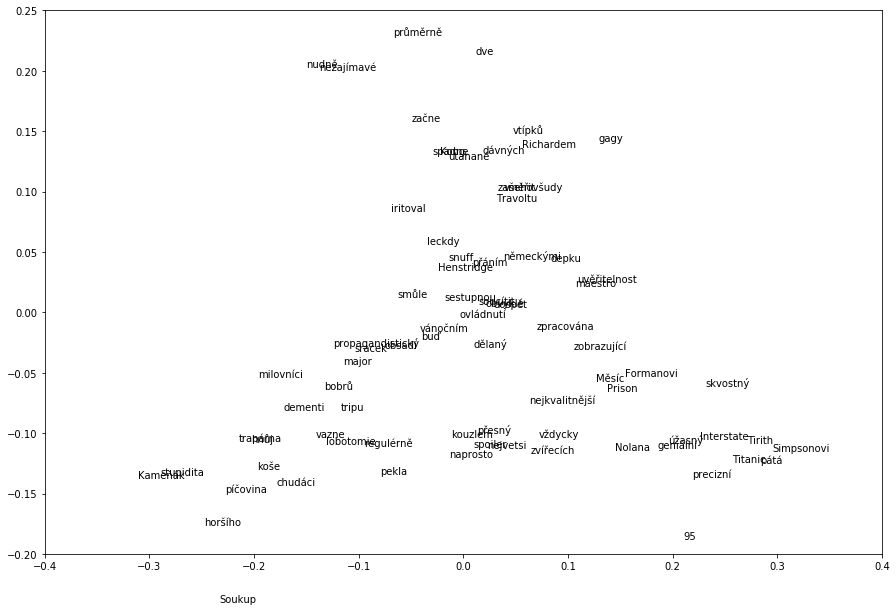}
  \put(70,38.45){\includegraphics[width=.31\textwidth]{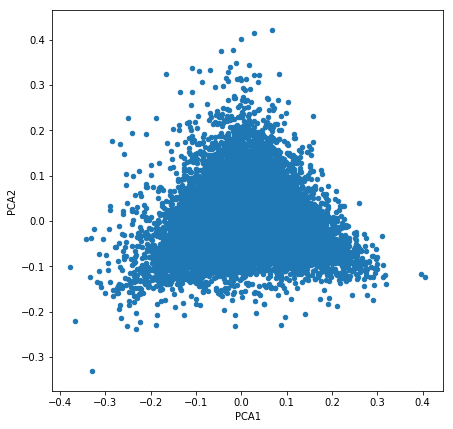}}
  \end{overpic}
    \caption{A random sample of words from the distribution of the embeddings from the sentiment analysis CNN model along the first (horizontal) and second (vertical) PCA dimension. The top right subplot shows the complete distribution.}
    \label{fig:emo_sample}
\end{figure}

The subplot in the top right of Figure~\ref{fig:emo_sample} shows the distribution of the embbedings projected on the first two dimensions of the PCA. They form a triangular shape which is unlikely to be formed by a random distribution.

To examine the distribution further, we plotted a sample of the words in Figure~\ref{fig:emo_sample}. The plot shows that in the bottom right corner there are words like “skvostný” (\emph{magnificent}), “precizní” (\emph{precise}), “geniální” (\emph{brilliant}) and also “Titanic” and “Simpsonovi” (\emph{the Simpsons}). These are positive words or film works that are generally regarded as epitome of quality (indicated also by their high ranking in the ČSFD database).  There is also the number “95” in the bottom right. The perceived positiveness of this number is due to users expressing their rating of the film as percentage in the comments.
In the bottom left there are words such as “stupidita” (\emph{stupidity}), “horšího” (\emph{worse}), “píčovina” (obscene word sometimes translated as \emph{bullshit}) and “Kameňák” (a Czech comedy). These are obviously negative words, with Kameňák being a film which while being relatively well known has a low score in the database (29\,\%), making it a suitable film to use as a negative comparison. This suggests that the first dimension of the PCA corresponds to valence or polarity of the word.

In the top of the figure, we see words such as “průměrné” (\emph{average, mediocre}) and “nezajímavé” (\emph{uninteresting}). These are neutral and not very intense. On the opposite side, in the middle near the bottom, we have words like “naprosto” (\emph{absolutely, completely}) and “největší” (\emph{biggest}) that are also neutral by themselves, but intensify the next word. Therefore, the second dimension corresponds to intensity. This also explains the triangular shape of the distribution: the words that are highly positive or highly negative are also intense, therefore there are words far right and left in the bottom part of the figure and not in the top part.


This is a stronger result than just being able to predict the emotional characteristics of the word from its embedding, because it shows that the embedding space of the sentiment analysis model is in the first place organised by these characteristics.

\section{Conclusion}
\label{sec:concl}

We examined the structure of Czech word embedding spaces with respect to POS and sentiment information.
Our method is based on correlating PCA dimension to categorical linguistic data.

We have demonstrated that although it is possible to successfully predict POS tags from embedding from various translation models and word2vec, not all of them show the same structure. Even though the information about POS is present in word2vec embeddings, the high degree of organization by POS in the NMT decoder suggests, that this information is important for machine translation and therefore the NMT model represents it in more direct way.

We have demonstrated that examining histograms of classes along the principal component is important to understand the structure of representation of information in embeddings. We have also demonstrated that NMT models represent named entities separately in the embedding space, word2vec distinguishes infinitive forms and modal verbs from the rest of the verbs, and CNN sentiment analysis models emotional properties of words in the shape of the embedding space. This shows that the embedding space of a neural network is literally shaped by the task for which it is trained.

The presented method of examining the representation can be extended to other languages, tasks and linguistic and non-linguistic information and may tell us a lot about the way that neural networks in NLP represent language. 

\section*{Acknowledments}

This work has been supported by the grant 18-02196S of the Czech
Science Foundation.
This research was partially supported by SVV
project number 260 453.

\bibliographystyle{splncs03}
\bibliography{paper}

\end{document}